\documentclass[letterpaper]{article} 
\usepackage{aaai19}  
\usepackage{times}  
\usepackage{helvet}  
\usepackage{courier}  
\usepackage{url}  
\usepackage{graphicx}  
\usepackage{amsmath}
\usepackage{amsfonts}
\usepackage{algpseudocode, algorithmicx, algorithm}
\usepackage{array}
\usepackage{soul}
\usepackage{booktabs}
\usepackage{bigstrut}
\newcolumntype{L}[1]{>{\raggedright\let\newline\\\arraybackslash\hspace{0pt}}m{#1}}
\newcolumntype{C}[1]{>{\centering\let\newline\\\arraybackslash\hspace{0pt}}m{#1}}
\newcolumntype{R}[1]{>{\raggedleft\let\newline\\\arraybackslash\hspace{0pt}}m{#1}}

\title{DeepChannel: Salience Estimation by Contrastive Learning \\for Extractive Document Summarization}
\author{Jiaxin Shi,\textsuperscript{1}\protect \thanks{Equal contribution.}
Chen Liang,\textsuperscript{1}\footnotemark[1]
Lei Hou,\textsuperscript{1}\protect \thanks{Corresponding Author.}
Juanzi Li,\textsuperscript{1}
Zhiyuan Liu,\textsuperscript{1}
Hanwang Zhang\textsuperscript{2}\\
\textsuperscript{1}{Tsinghua University}\\
\textsuperscript{2}{Nanyang Technological University}\\
\{shijx12,lliangchenc\}@gmail.com,
\{houlei,lijuanzi,liuzy\}@tsinghua.edu.cn,
hanwangzhang@ntu.edu.sg
}

\begin{document}
\maketitle
\begin{abstract}
We propose DeepChannel, a robust, data-efficient, and interpretable neural model for extractive document summarization. 
Given any document-summary pair, we estimate a salience score, which is modeled using an attention-based deep neural network, to represent the salience degree of the summary for yielding the document.
We devise a contrastive training strategy to learn the salience estimation network, and then use the learned salience score as a guide and iteratively extract the most salient sentences from the document as our generated summary.
In experiments, our model not only achieves state-of-the-art ROUGE scores on CNN/Daily Mail dataset, but also shows strong robustness in the out-of-domain test on DUC2007 test set. 
Moreover, our model reaches a ROUGE-1 F-1 score of 39.41 on CNN/Daily Mail test set with merely $1 / 100$ training set, demonstrating a tremendous data efficiency. 
\end{abstract}

\section{Introduction}

Automatic document summarization is a challenging task in natural language understanding, aiming to compress a textual document to a shorter highlight that contains the most representative information of the original text.
Existing summarization approaches are mainly classified into two categories: extractive methods and abstractive methods.
Extractive summarization methods, on which this paper focuses, aim to select salient snippets, sentences or passages directly from the input document, while abstractive summarization generates summaries that may have words or phrases not present in the input.

Recently, as end-to-end deep learning has made great progress in many NLP fields, such as machine translation~\cite{luong2015effective} and question answering~\cite{iyyer2014neural}, a lot of researchers have proposed neural models to address the document summarization problem.
For example, SummaRuNNer~\cite{nallapati2017summarunner} uses a Recurrent Neural Network (RNN) based sequence model for extractive summarization, Refresh~\cite{narayan2018ranking} assigns each document sentence a score to indicate its probability of being extracted, and many abstractive models~\cite{see2017pointer,jadhav2018extractive} are developed based on the encoder-decoder framework that encodes a document and decodes its summary.
These existing neural summarizers mostly aim to build an end-to-end mapping from the input document to its summary.
The learning of such an end-to-end neural network 1) always requires a huge amount of training corpus, 2) easily suffers from the overfitting problem~\cite{srivastava2014dropout,erhan2010does}, and 3) usually lacks interpretability.

\begin{table}[t]
  \small
  \centering
    \begin{tabular}{|p{0.9\linewidth}|}
    \toprule
    $D$:  Rutgers University has banned fraternity and sorority house parties at its main campus in New Brunswick, New Jersey, for the rest of the spring semester after several alcohol-related problems this school year, including the death of a student. \\
    \midrule
    $S_1$: Rutgers University has banned fraternity and sorority house parties because of an alcohol-related accident that led to the death of a student. \\
    \midrule
    $S_2$: The main campus of Rutgers University is located in New Brunswick, New Jersey. \\
    \bottomrule
    \end{tabular}%
  \caption{Examples of different degrees of salience. We consider $P(D|S_1) > P(D|S_2)$ because $S_1$ contains more important information compared with $S_2$ and thus is more salient for yielding $D$.  \label{tab:example}}
\end{table}%

To alleviate these problems, we propose a neural extractive summarizer named \textbf{DeepChannel}, which estimates salience for guiding the extraction procedure instead of learning an end-to-end mapping.
DeepChannel is inspired by the noisy-channel~\cite{knight2002summarization,daume2002noisy}, a probabilistic approach for sentence-level and document-level compression.
Given an input document $D$, the noisy-channel model aims to find an optimal summary $S$ that maximizes $P(S|D)$.
It 1) splits $P(S|D)$ using Bayes rule, 2) independently estimates a language model probability $P(S)$ and a channel model probability $P(D|S)$, 3) defines expanding rules, and 4) learns the parameters in a traditional statistical manner.
Such a statistical approach depends on manual rules, lacks generality, suffers from data sparsity, and fails to capture semantics~\cite{hinton2009LM}, which is the key for document understanding.
To this end, we design a neural channel model to draw support from the great representation power of deep learning.

Given any document-summary pair $(D, S)$, we learn a channel probability (i.e., salience score) $P(D|S)$, representing that we start with a short summary $S$ and add ``noise'' to it, yielding a longer document, how likely $D$ is produced.
It can be considered as a measure of how much salient information of $D$ is contained in $S$.
Table~\ref{tab:example} gives an example where $S_1$ is more salient than $S_2$ for yielding $D$.
We design an attention-based neural network to model the channel probability, and train it with a contrastive training strategy.
That is, we firstly use a heuristic way to randomly produce contrastive samples, including two candidate summaries $S_1$ and $S_2$ for an input $D$ where the former is more salient, and then maximize the margin between $P(D|S_1)$ and $P(D|S_2)$.
This training strategy implicitly increases the size of training instances and incorporate randomness into the training procedure, and thus help our model perform well even on a small training set.
With a well-learned $P(D|S)$, we produce the optimal summary $S^* = \text{argmax}_{S} P(D|S)$ by greedily extracting the most salient sentences which have a maximum probability to expand to the whole document.
Compared with the statistical noisy-channel, our neural model can 1) make use of semantics involved in distributed representations, 2) alleviate the training sparseness and 3) avoid the high-cost expert-designed rules.
\footnote{$P(S)$ is not taken into consideration in our current model, and we leave it for future work.}

Our model consists of two parts, salience estimation and salience-guided extraction.
Only the first part is parametric and requires an annotated corpus for training.
Different from most state-of-the-art approaches that usually learn a direct mapping from a document to its annotated summary, our salience estimation learns a mapping from any document-summary pair to a salience score.
It brings two significant benefits:
1) Our model is more robust to domain variations.
DeepChannel performs much better than other end-to-end baselines when testing on DUC 2007~\footnote{\url{https://www-nlpir.nist.gov/projects/duc/guidelines/2007.html}} while training on CNN/Daily Mail~\footnote{\url{https://github.com/deepmind/rc-data}}.
2) Our model is much more data-efficient and alleviates the overfitting problem to a great degree. 
DeepChannel performs well even when we reduce the size of the CNN/Daily Mail training set to $1/100$.

We also conduct quantitative and qualitative experiments on the standard CNN/Daily Mail benchmark, demonstrating that our model not only performs on par with state-of-the-art summarization systems, but also shows high interpretability due to the well-designed attention mechanism.

To sum up, our contributions are as follows:
\begin{itemize}
    \item we propose DeepChannel, an extractive summarization approach consisting of a deep neural network for salience estimation and a salience-guided greedy extraction strategy;
    \item we demonstrate that our model outperforms or matches state-of-the-art summarizers, is robust to domain variations, performs well on the small training set, and is highly interpretable.
\end{itemize}

\section{Related Work}

Traditional summarization methods usually depend on manual rules and expert knowledge, such as the expanding rules of noisy-channel models~\cite{daume2002noisy,knight2002summarization}, objectives and constraints of Integer Linear Programming (ILP) models~\cite{woodsend2012multiple,parveen2015topical,bing2015abstractive}, human-engineered features of some sequence classification methods~\cite{shen2007document}, and so on.

Deep learning models can learn continuous features automatically and have made substantial progress in multiple NLP areas.
Many deep learning-based summarization models have been proposed recently for both extractive and abstractive summarization tasks.

\textbf{Extractive.} 
\cite{nallapati2017summarunner} considers the extraction as a sequence classification task and proposes SummaRuNNer, a simple RNN based model that decides whether or not to include a sentence in the summary.
\cite{wu2018learning} takes the coherence of summaries into account and designs a reinforcement learning (RL) method to maximize the combined ROUGE~\cite{lin2004rouge} and coherence reward.
\cite{narayan2018ranking} conceptualizes extractive summarization as a sentence ranking task and optimizes the ROUGE evaluation metric through an RL objective.
\cite{jadhav2018extractive} models the interaction of keywords and salient sentences using a two-level pointer network and combines them to generate the extractive summary.

\textbf{Abstractive.}
A vast majority of abstractive summarizers are built based on the encoder-decoder structure.
\cite{see2017pointer} incorporates a pointing mechanism into the encoder-decoder, such that their model can directly copy words from the source text while decoding summaries.
\cite{paulus2017deep} combines the standard cross-entropy loss and RL objectives to maximize the ROUGE metric at the same time of sequence prediction training.
\cite{chen2018fast} proposes a fast summarization model that first selects salient sentences and then rewrites them abstractively to generate a concise overall summary.
Their hybrid approach jointly learns an extractor and a rewriter, capable of both extractive and abstractive summarization.
\cite{hsu2018unified} also combines extraction and abstraction, but they implement it by unifying a sentence-level attention and a word-level attention and guiding these two parts with an inconsistency loss.

Most of these deep summarization models aim to learn a direct mapping from the document to the summary.
Instead, our DeepChannel aims to learn a channel probability to measure the salience of any document-summary pair.
\cite{peyrard2017supervised} learns to estimate automatic Pyramid scores and extract summaries by solving an ILP problem, but their model depends on a lot of manual features and their ILP-based extraction is totally different from ours.

\section{DeepChannel}

\begin{figure*}[ht]
\includegraphics[width=\linewidth]{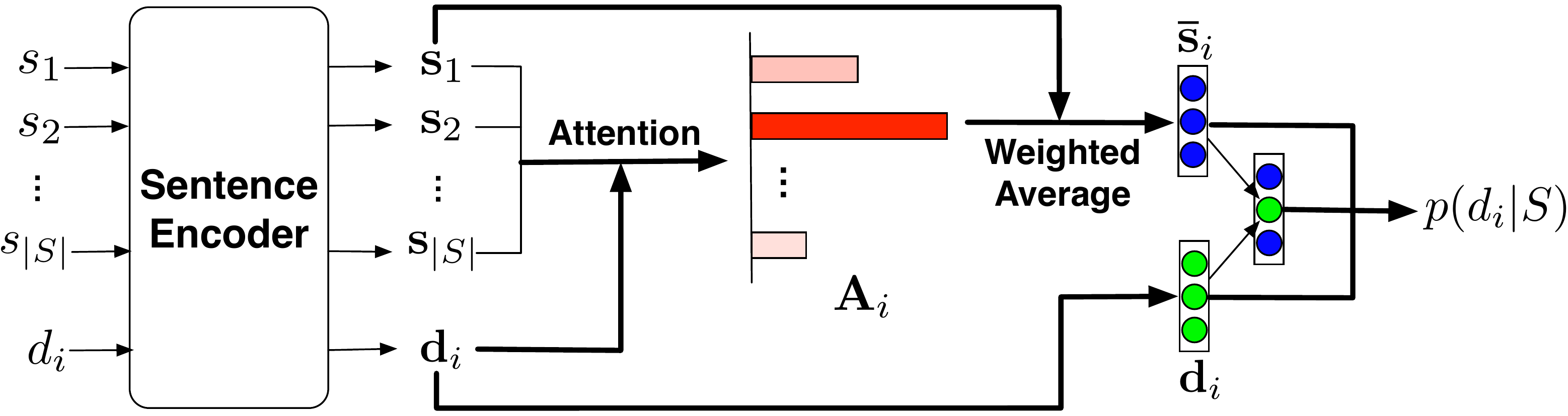}
\caption{Our attention-based channel model to compute $P(d_i|S)$. After encoding all sentences into dense vectors, we regard $\mathbf{d}_i$ as a query and assign an attention weight to each summary sentence. Then we combine $\mathbf{d}_i$ and the weighted average $\mathbf{\bar{s}}_i$ together to compute the channel probability $P(d_i|S)$.}
\label{fig:model}
\end{figure*}

We represent a document-summary pair as $(D, S)$, where $S$ is an either annotated or generated summary for document $D$.
$D$ consists of $|D|$ sentences $[d_1, d_2, \cdots, d_{|D|}]$, and $S$ consists of $|S|$ sentences $[s_1, s_2, \cdots, s_{|S|}]$.
The $i\text{-th}$ sentence of the document can be represented as a sequence of words, i.e., $d_i = [dw_{i,1}, dw_{i,2}, \cdots, dw_{i,|d_i|}]$, where $dw_{i,j}$ denotes the $j\text{-th}$ word of $d_i$.
Similarly, we have $s_i = [sw_{i,1}, sw_{i,2}, \cdots, sw_{i,|s_i|}]$.

Our DeepChannel model consists of two parts.
The first is an attention-based neural network for \emph{salience estimation}, which takes $(D, S)$ as input and outputs the channel probability $P(D|S)$, representing the chance that $D$ is generated in terms of $S$.
This part is trained using a novel contrastive training strategy and then serves for the extraction.
The second is a greedy extraction strategy, which utilizes the learned salience estimation model to extract the most salient sentences from the original document.
We denote the golden summary that is annotated in the training corpus as $\hat{S}$, and our extracted summary as $S^*$.

\subsection{Neural Salience Estimation}
For estimating $P(D|S)$, we consider that the document $D$ is generated based on the given $S$.
For simplicity, we assume that sentences in the document are conditional independent. 
Then we have $P(D|S) = \prod\nolimits{_{i=1}^{|D|}} P(d_i|S)$, where $P(d_i|S)$ denotes the chance that $d_i$ is produced from $S$.
Another assumption is that different summary sentences make different amounts of contribution to the generation of $d_i$.
When calculating $P(d_i|S)$, we should concentrate more on those summary sentences that have higher semantic relevance to $d_i$.
We use an attention mechanism to model this.

As our target is the probability value rather than to decode the texts, we compute the probability just in sentence-level instead of further deriving the equation to a word-level sequence generation process (i.e., the encoder-decoder).
Some sentence embedding models~\cite{quickthoughts} use the similar simplification strategy, which makes the learning much more efficient.

Specifically, we encode each sentence of $(D, S)$ via a Gated Recurrent Unit (GRU)~\cite{chung2014empirical}, one of the most renowned variants of RNNs, to obtain the sentence-level semantic vectors:

\begin{equation}
\begin{aligned}
\mathbf{d}_i &= \text{GRU}([dw_{i,1}, \cdots, dw_{i,|d_i|}]; \theta_1), &i=1,\cdots,|D|\\
\mathbf{s}_j &= \text{GRU}([sw_{j,1}, \cdots, sw_{j,|s_j|}]; \theta_1), &j=1,\cdots,|S|.
\end{aligned}
\end{equation}
Sentences of the document and the summary share the same encoder whose parameters are denoted as $\theta_1$.

To compute $P(d_i|S)$, we design an attention mechanism (see Figure~\ref{fig:model}) that assigns a weight $\mathbf{A}_{i,j}$ to each summary sentence $s_j$, which will be large if the semantics of $s_j$ is similar to $d_i$.
Then we calculate the weighted summation of summary sentence vectors, denoted by $\mathbf{\bar{s}}_i$, concatenate it with $\mathbf{d}_i$, and feed them into a multi-layer perceptron (MLP).
Besides, for further information interaction, we take the element-wise production of these two vectors, $\mathbf{d}_i \odot \mathbf{\bar{s}}_i$, as another input of MLP.
Formally, we have

\begin{equation}\label{for:channel}
\begin{aligned}
\mathbf{A}_i &= \text{Softmax}(\mathbf{d}_i^{\top} [\mathbf{s}_1; \mathbf{s}_2; \cdots; \mathbf{s}_{|S|}]),\\
\mathbf{\bar{s}}_i &= \sum\limits_{j=1}^{|S|} \mathbf{A}_{i,j} \mathbf{s}_j,\\
P(d_i|S) &= \text{Sigmoid}(\text{MLP}([\mathbf{d}_i; \mathbf{\bar{s}}_i; \mathbf{d}_i \odot \mathbf{\bar{s}}_i]; \theta_2)),\\
\end{aligned}
\end{equation}

where $\theta_2$ is parameters of MLP.
Let $\theta$ include both $\theta_1$ and $\theta_2$, we can reformulate our channel probability as
\begin{equation}\label{for:PDS}
P(D|S; \theta) = \prod\limits_{i=1}^{|D|} \text{Sigmoid}(\text{MLP}([\mathbf{d}_i; \mathbf{\bar{s}}_i; \mathbf{d}_i \odot \mathbf{\bar{s}}_i])).
\end{equation}

\subsection{Contrastive Learning}

We expect that $P(D|S)$ should be large if $S$ contains salient information to construct $D$, else it should be small.
To achieve this goal, we devise a contrastive training strategy.
That is, given a document $D$, we construct a pair of contrastive candidate summaries $S_1$ and $S_2$, one positive and one negative, satisfying that $S_1$ is more salient to summarize $D$ than $S_2$.
Then we train our channel model by maximizing the margin between $P(D|S_1)$ and $P(D|S_2)$.

Given an annotated pair $(D, \hat{S})$, we consider that the golden summary $\hat{S}$ is salient to construct $D$, and all summary sentences contain the necessary information.
Therefore, we can assume that when we delete a sentence from $\hat{S}$ or replace it with another sentence which has different meanings, some key information will lose and the salience score is expected to drop.

\begin{figure}[t]
\includegraphics[width=\linewidth]{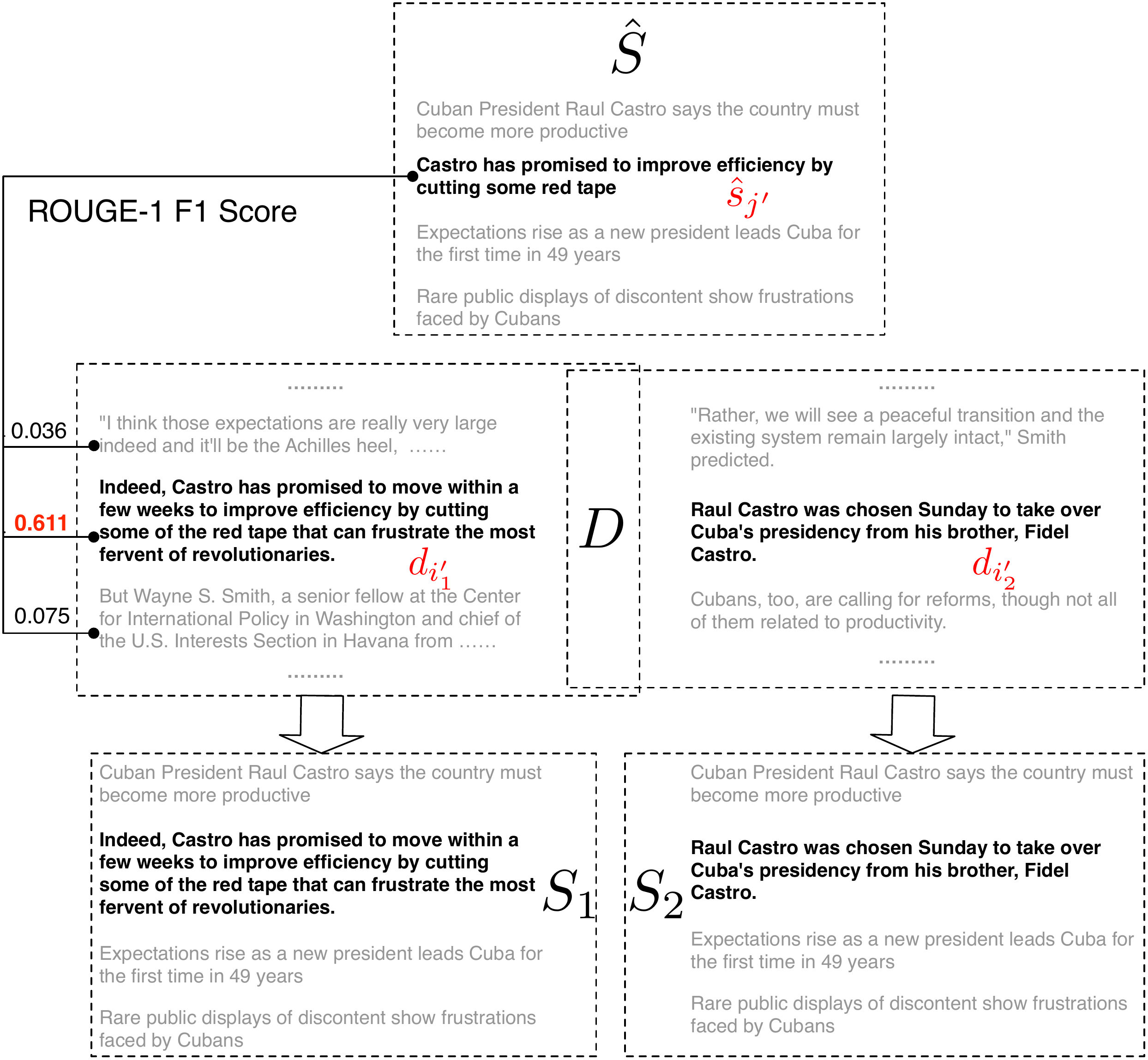}
\caption{An instance of our contrastive samples. Given an annotated $(D, \hat{S})$, we randomly discard a summary sentence $\hat{s}_{j'}$, and fill $d_{i'_1}$ and $d_{i'_2}$ to form the contrastive pair. $d_{i'_1}$ has the highest ROUGE score with $\hat{s}_{j'}$. $d_{i'_2}$ is randomly sampled.}
\label{fig:contrastive}
\end{figure}

Based on this assumption, we construct the negative candidate $S_2$ by randomly selecting one sentence $\hat{s}_{j'}$ from the golden summary and then replacing it with a randomly selected document sentence $d_{i'_2}$.
$\hat{S}$ is a straightforward positive candidate, but it will cause that all positive sentences are from the summary space while the negative candidate contains document sentences.
Such an information asymmetry may mislead the learning process.
To this end, we obtain $S_1$ by calculating the ROUGE~\cite{lin2004rouge} scores, specifically, the ROUGE-1 F-1 scores, between each document sentence and the discarded $\hat{s}_{j'}$, and then using $d_{i'_1}$ which has the highest ROUGE with $\hat{s}_{j'}$ to replace it.
Figure~\ref{fig:contrastive} gives an example.

As ROUGE is a metric of sentence similarity, we can consider that $d_{i'_1}$ contains similar information to $\hat{s}_{j'}$.
We do not use the document sentence with the minimum ROUGE score in the negative candidate.
Random sampling can not only strengthen the robustness, but also increase the difficulty of distinguishing the contrastive pair, which may provide sharper training signals.

The loss function of contrastive training can be formulated as
\begin{equation}
\mathcal{L}_{con}(\theta) = -\big( \log P(D|S_1) - \log P(D|S_2) \big)
\end{equation}
for each $(D,\hat{S})$ of training corpus.

\subsection{Penalization Term}

Let $\mathbf{A}$ denote the attention matrix of $(D,S)$.
We consider that a reasonable attention should satisfy following two conditions:
1) $\mathbf{A}_i$ is sharp, that is, the $i$-th document sentence should focus on its most relevant summary sentences.
2) All summary sentences are important and each summary sentence should get attention from some document sentences.
Inspired by \cite{lin2017structured}, we introduce a penalization term to achieve both of these two goals:

\begin{equation}
\mathcal{L}_{penal}(\theta) = \big|\big| \mathbf{A}^\top \mathbf{A} - \frac{|D|}{|S|}\mathbf{I} \big|\big|_F.
\end{equation}

Here $||\cdot||_F$ stands for the Frobenius norm of a matrix, the shape of $\mathbf{A}$ is $|D| \times |S|$ and the shape of $\mathbf{I}$ is $|S| \times |S|$. This penalization term will be minimized together with the contrastive loss.

Because of the softmax, we have $\sum_{i=1}^{|S|} \mathbf{A}_{k,i} = 1$ for any valid $k$.
We denote the element in the $\mathbf{A}^\top \mathbf{A}$ matrix as $a_{i,j}$, which is equal to the inner production of $\mathbf{A}_{:,i}$ and $\mathbf{A}_{:,j}$.
As all elements in $\mathbf{A}$ is non-negative, we can draw that 1) $a_{i,j} \ge 0$, and 2) $a_{i,j}=0$ \textit{iff.} $\mathbf{A}_{k,i}=0$ or $\mathbf{A}_{k,j}=0$ for any $k$.
In other words, if $\mathbf{A}_k$ is not sharp and attends to $s_i$ and $s_j$ at the same time, then $a_{i,j}$ will be greater than $0$.
By forcing the non-diagonal $a_{i,j} (i \ne j)$ to approximate $0$, we can encourage each $d_k$ to focus on summary sentences as sharp as possible.
On the other hand, we force the diagonal of $\mathbf{A}^\top \mathbf{A}$ to approximate $\frac{|D|}{|S|}$, meaning that each summary sentence should receive nearly average attention, avoiding that certain $s$ is not focused on at all.
To understand this intuitively, let's consider that each row in $\mathbf{A}$ is a one-hot vector, meaning that each document sentence attends to only one summary sentence.
Then $a_{i,i}$ is totally equal to the number of received attention of $s_i$, and $\sum_{i=1}^{|S|} a_{i,i} = |D|$.
The diagonal part of our penalization term amounts to encouraging an average division of these attention.
This simple average assumption is not accurate but is efficient to compute and is demonstrated to be effective.

Our final loss function is:
\begin{equation}
\mathcal{L}(\theta) = \mathcal{L}_{con} + \alpha \mathcal{L}_{penal},
\end{equation}
where $\alpha$ is a hyperparameter, and $\mathcal{L}_{penal}$ is computed using $(D, S_1)$.

\subsection{Greedy Extraction}
For testing, we devise a greedy extraction strategy in terms of our well-trained channel model $P(D|S)$, described in Algorithm~\ref{algo:greedy}.

\begin{algorithm}
\caption{Greedy Extraction Algorithm\label{algo:greedy}}
\textbf{Input:} document $D = \{ d_1, d_2, ..., d_{|D|} \} $ , a well-pretrained channel model $P(D|S)$, expected summary length $l$\\
\textbf{Output:} optimal summary $S^*$
\begin{algorithmic}
    \State $S^* \gets \{\}$
    \While{$|S^*| < l$}

        \State $d,\ \ p \gets nil,\ \ 0$
        \For{$d_i \in D - S^*$}
            \State $p_i \gets P(D|S^* \cup \{d_i\})$ \text{according to Formula~\ref{for:PDS}}
            \If{$p_i > p$}
                \State $d,\ \ p = d_i,\ \ p_i$
            \EndIf
        \EndFor
        \State $S^* = S^* \cup \{d\}$
    \EndWhile
    \State \text{Resort $S^*$ based on the order in $D$}
    \State \Return $S^*$
\end{algorithmic}
\end{algorithm}

We iteratively extract one sentence from the document and add it into $S^*$, such that $P(D|S^*)$ is greedily maximized until the upper bound of the length of the summary $l$ is reached.
Such a simple greedy extraction algorithm is computationally efficient.
Furthermore, it can automatically avoid redundancy between extracted sentences, because the salience score of $S^*$ will not increase if we add a redundant sentence into $S^*$.
Benefiting from the great potential of the channel model, what we extract at each step must be unique and valuable.
We will further demonstrate that in our experiments.

In Algorithm~\ref{algo:greedy} we exclude $S^*$ from $D$ at each step because we observed some ``magic sentence''s in experiments.
That is, after a document sentence $d_{magic}$ is extracted into $S^*$, appending any other $d_i, i \ne magic$ into $S^*$ will lead to a decrease of $P(D|S^*)$, and thus $d_{magic}$ will be repeatedly selected as it can hold that probability. 
We guess it is because $d_{magic}$ is much more salient than other $d_i$, and appending other $d_i$ into $S^*$ will ``distract'' the channel attention.

Using this greedy extraction strategy, we can produce an extracted summary containing $l$ sentences for any given input document.

\section{Experiments}

\subsection{Datasets}
We evaluate our model on two datasets: CNN/Daily Mail~\cite{hermann2015teaching,nallapati2016abstractive,see2017pointer,hsu2018unified} and DUC 2007.
The CNN/Daily Mail dataset contains news stories in CNN and Daily Mail websites and corresponding human-written highlights as summaries.
This dataset has two versions: anonymized, which replaces named entities by special tokens, and non-anonymized, which preserves the raw texts.
We follow \cite{hsu2018unified} and obtain the non-anonymized version of this dataset which has 287,113 training pairs, 13,368 validation pairs, and 11,490 test pairs.

DUC 2007 is a multiple-document dataset containing 45 topics, and each topic corresponds to 25 relevant documents and 4 summary annotations.
We concatenate multiple documents in the same topic to obtain a single-document test set whose size is 45.
After training on CNN/Daily Mail, we use DUC 2007 dataset as an additional out-of-domain test set, to compare the robustness of different models.

\subsection{Implementation Details}
For preprocessing, we lower all document and summary sentences, replace numbers with a placeholder ``$\langle$zero$\rangle$'' and remove sentences containing less than 4 words.
We set the vocabulary size to 50k and replace low-frequency words with a special token ``$\langle$unk$\rangle$''.

For the model, we set the dimension of the word embedding to 300, and the GRU hidden dimension to 1024.
We use a 3-layered MLP to calculate $P(d_i|S)$ in Formula~\ref{for:channel}, which consists of 3 linear layers, 2 ReLU layers and an output sigmoid layer.
We use dropout~\cite{srivastava2014dropout} with probability 0.3 after the word embedding layer and before the first layer of the MLP.

For the training and hyperparameters, we init our word embeddings using GloVe~\cite{pennington2014glove} pretrained vectors and then finetune them in our task.
We use Adam~\cite{kingma2014adam} optimizer with a fixed learning rate of 1e-5 to train our model.
We set the weight of the penalization term $\alpha=0.001$.
When extracting sentences, we fix the number of target sentences (i.e., $l$ in Algorithm~\ref{algo:greedy}) to 3.
The implementation is made publicly available.\footnote{\url{https://github.com/lliangchenc/DeepChannel}}

\subsection{Evaluation}
For CNN/Daily Mail experiments, we use the full-length Rouge F1 metric~\cite{lin2004rouge}.
For DUC 2007, we use limited length Rouge recall at 75 bytes and 275 bytes.
We report the scores from Rouge-1, Rouge-2, and Rouge-L, which are computed using the matches of unigrams, bigrams, and longest common subsequences respectively, with the ground truth summaries.

\subsection{Baselines}
Our extractive baselines include: lead-3~\cite{see2017pointer}, SummaRuNNer~\cite{nallapati2017summarunner}, Refresh~\cite{narayan2018ranking}, SWAP-NET~\cite{jadhav2018extractive}, and rnn-ext+RL~\cite{chen2018fast}.

We also compare our performance with state-of-the-art abstractive baselines, including PointerGenerator~\cite{see2017pointer}, ML+RL+intra-attention~\cite{paulus2017deep}, controlled~\cite{fan2017controllable}, and inconsistency loss~\cite{hsu2018unified}.

For further analyses such as out-of-domain test, we select the 3 most representative approaches, SummaRuNNer, Refresh, and PointerGenerator , as the baselines.
SummaRuNNer predicts a binary label for each document sentence, indicating whether it is extracted.
Refresh learns to rank sentences using reinforcement learning and then directly extracts the top-$k$.
PointerGenerator, which is built on the sequence-to-sequence (seq2seq) framework, is one of the most typical abstractive summarizers.

\section{Results}

\subsection{Results on CNN/Daily Mail}

Table~\ref{tab:cnndaily} shows the performance comparison between our DeepChannel and state-of-the-art baselines on the CNN/Daily Mail dataset using full-length Rouge F-1 as the metric.
We can see that DeepChannel performs better than or at least on par with state-of-the-art models.
Besides DeepChannel, there are two approaches achieving more than 41.0 Rouge-1 scores: SWAP-NET and rnn-ext + RL.
SWAP-NET combines the word-level extraction and the salient sentence selection, such a fine-grained extraction brings it great performance gain.
The other one, rnn-ext + RL, benefits from directly regarding Rouge scores as reward signals in the reinforcement learning.
Our model has much simpler structures than them but can still achieve comparable performance.
Moreover, due to the simple structure, our model converges very fast. 
To obtain the results in Table~\ref{tab:cnndaily}, DeepChannel only needs to be trained one epoch on CNN/Daily Mail training set, taking about four hours with an Nvidia GTX 1080Ti GPU.

\begin{table}[htbp]
  \small
  \centering
    \begin{tabular}{|l|c|c|c|}
    \hline
    Method & Rouge-1 & Rouge-2 & Rouge-L \\
    \hline
    \multicolumn{4}{|l|}{Extractive} \\
    \hline
    lead-3 & 40.34 & 17.70  & 36.57 \\
    SummaRuNNer & 39.60  & 16.20  & 35.30 \\
    Refresh & 40.00    & 18.20  & 36.60 \\
    SWAP-NET & 41.60  & 18.30  & 37.70 \\
    rnn-ext + RL & 41.47 & 18.72 & 37.76 \\
    DeepChannel  &    41.50   &    17.77   &  37.62 \\
    \hline
    \multicolumn{4}{|l|}{Abstractive} \\
    \hline
    PointerGenerator & 39.53 & 17.28 & 36.38 \\
    ML+RL+intra-attention & 39.87 & 15.82 & 36.90 \\
    controlled & 39.75 & 17.29 & 36.54 \\
    inconsistency loss & 40.68 & 17.97 & 37.13 \\
    \hline
    \end{tabular}%
  \caption{Performance on CNN/Daily Mail test set using the full length Rouge F-1 score.\label{tab:cnndaily}}
\end{table}%

\subsection{Results on DUC 2007}

\begin{table}[htbp]
  \small
  \centering
    \begin{tabular}{|c|c|c|c|}
    \hline
          & Rouge-1 & Rouge-2 & Rouge-L \\
    \hline
    \multicolumn{4}{|c|}{75 bytes} \\
    \hline
    SummaRuNNer & 18.32 & 4.57  & 12.96 \\
    PointerGenerator & 13.74 & 2.49  & 10.97 \\
    Refresh & 18.39 & 5.04  & 14.85 \\
    DeepChannel  & \textbf{19.53} & \textbf{5.12}  & \textbf{15.88} \\
    \hline
    \multicolumn{4}{|c|}{275 bytes} \\
    \hline
    SummaRuNNer & 27.06 & 6.09  & 6.49 \\
    PointerGenerator & 23.93 & 4.70   & 5.98 \\
    Refresh & 26.80  & 6.30   & 6.66 \\
    DeepChannel  & \textbf{28.85} & \textbf{6.86}  & \textbf{6.80} \\
    \hline
    \end{tabular}%
  \caption{Performance on DUC 2007 dataset using the \textbf{limited length recall} variants of Rouge. The upper section are results at 75 bytes, and the lower are results at 275 bytes. DeepChannel outperforms other baselines stably, indicating that it is more robust for the out-of-domain application.\label{tab:duc}}
\end{table}%

To compare the robustness of models, we conducted out-of-domain experiments by training models on CNN/Daily Mail training set while evaluating on DUC 2007 dataset.
Table~\ref{tab:duc} shows the limited length Rouge recall scores at 75 bytes and 275 bytes.
We can see that DeepChannel obtains Rouge-1 score of 19.53 at 75 bytes and 28.85 at 275 bytes, stably and significantly better than other three baselines, demonstrating the strong robustness of our model.

It is worth noting that PointerGenerator, a seq2seq based abstractive approach, suffers performance drop by a large margin when transferred to the out-of-domain dataset.
After being trained on CNN/Daily Mail training set, it performs on par with SummaRuNNer and Refresh when testing on CNN/Daily Mail test set (Table~\ref{tab:cnndaily}), while worse a lot on DUC 2007.
We consider that the seq2seq summarization systems are more easily to suffer from the overfitting problem as they attempt to memorize as many details (i.e., learn to decode each word) of the training data as possible.

\subsection{Results on Reduced CNN/Daily Mail}

\begin{table}[htbp]
  \small
  \centering
    \begin{tabular}{|c|c|c|c|}
    \hline
          & Rouge-1 & Rouge-2 & Rouge-L \\
    \hline
    \multicolumn{4}{|c|}{\textbf{1/10 (28,711 training samples)}} \\
    \hline
    SummaRunner &  35.95 &  15.87  &  32.38 \\
    PointerGenerator & 34.32 & 11.82 & 31.54 \\
    Refresh & 36.30  & 14.56 & 33.06 \\
    DeepChannel & \textbf{40.49} & \textbf{17.07} &  \textbf{36.59}  \\
    \hline
    \multicolumn{4}{|c|}{\textbf{1/100 (2,871 training samples)}} \\
    \hline
    SummaRunner &   35.44 &  15.50  &  31.88   \\
    PointerGenerator & 28.57 & 6.28  & 25.90 \\
    Refresh & 36.05 & 14.23 & 32.79 \\
    DeepChannel  &  \textbf{39.41}  & \textbf{16.15} &  \textbf{35.64}  \\
    \hline
    \end{tabular}%
  \caption{Performance when training on reduced CNN/Daily Mail training set. The full-length Rouge F-1 scores on CNN/Daily Mail test set are reported. The two sections are results of 1/10 and 1/100 respectively. Our model can obtain high scores even with only 1/100 training samples, while other baselines, especially the seq2seq-based PointerGenerator, suffer a significant performance degradation on reduced training set. \label{tab:reduced}}%
\end{table}%

We reduced the size of the training set to explore the data efficiency of different models.
We conducted two experiments, respectively preserving 1/10 (28,711 pairs) and 1/100 (2,871 pairs) samples of the CNN/Daily Mail training set.

Models were trained on the reduced training set and evaluated on the original test set.
Table~\ref{tab:reduced} shows the performance of different models, using full-length Rouge F-1 as the measurement.

We can see that being trained on merely 2,871 training samples, our DeepChannel can still achieve a good Rouge score, just slightly lower than the score obtained on the complete training set.
In contrast, the Rouge score of SummaRunner, Refresh, and especially PointerGenerator, all suffer a drastic drop on the reduced training set.
When the fraction reduces from 1/10 to 1/100, PointerGenerator's Rouge-1 F1 score drops sharply, i.e., from 34.32 to 28.57.
We think it is due to the same reason as why PointerGenerator performs badly on DUC 2007.
The seq2seq structure attempts to learn all details of the training set, leading to a more serious overfitting problem when the number of training samples is limited.
Attributed to our salient estimation, DeepChannel has strong generalization ability and can learn from a very small training set and avoid overfitting to a great extent.

\begin{table}[htbp]
  \small
  \centering
    \begin{tabular}{|p{0.95\linewidth}|}
    \hline
    \textit{Document:} \textbf{Rutgers University has banned fraternity and sorority house parties at its main campus in New Brunswick, New Jersey, for the rest of the spring semester after several alcohol-related problems this school year, including the death of a student. The probation was decided last week but announced by the university Monday.} 'Rutgers takes seriously its ... university said in a statement. \textbf{Last month, a fraternity was shut down because of an underage drinking incident in November in which a member of Sigma Phi Epsilon was taken to a hospital after drinking heavily at the fraternity house.} Rutgers University has banned fraternity and sorority house parties at its main campus for the rest of the spring semester after several alcohol-related problems  ...... \\
    \hline
    \textit{Gold Summary:} \ul{Rutgers University has banned fraternity and sorority house parties at its main campus for the rest of the spring semester. The probation was decided last week, but the school announced the move on Monday.} 86 recognized fraternities and sororities will be allowed to hold spring formals and other events where third-party vendors serve alcohol. \ul{Last month, a fraternity was shut down because of an underage drinking incident in November. A member of Sigma Phi Epsilon was taken to a hospital after drinking heavily at the fraternity house during the incident.} In September, a 19-year-old student, Caitlyn Kovacs, died of alcohol poisoning after attending a fraternity party. \\

    \hline
    \hline
    \textit{Document:} ......  are not as kind on the body as they purport to be. \textbf{Investigators found that a number of flavors were labeled 'healthy' - brimming with fiber, protein and antioxidants, while being low in fat and sodium. However, upon closer inspection, it was found that 'none of the products met the requirements to make such content claims' and were in fact 'misbranded'.} Mislabeled? The FDA has ruled that KIND bars are not as kind on the body as they purport to be. \textbf{Indeed, Daily Mail Online calculated that one KIND bar flavor - not included in the FDA investigation - contains more calories, fat and sodium than a Snickers bar.} A 40g Honey Smoked BBQ KIND Bar ......\\
    \hline
    \textit{Gold Summary:} \ul{FDA Investigators found that a number of flavors were labeled 'healthy' - brimming with fiber and antioxidants, while being low in fat and sodium. However, upon closer inspection it was found that 'none of the products met the requirements to make such content claims'. Daily Mail Online calculated that one KIND bar flavor - not included in the FDA investigation - contains more calories and fat than a Snickers bar. } New York University nutritionist, Marion Nestle, likened KIND bars to candy \\
    \hline
    \end{tabular}%
  \caption{Example documents and gold summaries from CNN/Daily Mail test set. The sentences chosen by DeepChannel for extractive summarization are highlighted in bold, and the corresponding summary sentences which have equivalent semantics are underlined. \label{tab:qualitative}}
\end{table}%

\subsection{Influence of the Penalization Term}

\begin{table}[htbp]
\small
\centering
    \begin{tabular}{|c|c|c|c|}
    \hline
    $\alpha$      & Rouge-1 & Rouge-2 & Rouge-L \\
    \hline
    0 & 40.89  &     17.21   &     37.08 \\
    0.001 & \textbf{41.50}  &     \textbf{17.77}  &     \textbf{37.62} \\
    0.01 &  41.30  &     17.75  &     37.43 \\
    0.1  & 40.49  &     17.23  &     36.65 \\
    \hline
    \end{tabular}%
  \caption{Performance on CNN/Daily Mail test set with different weights of the penalization term. \label{tab:penal}}
\end{table}%

We set $\alpha$ --- the weight of the penalization term --- to $0.001$ in our experiments.
In Table~\ref{tab:penal} we show results of different $\alpha$ values, to illustrate why we choose $0.001$.
When we remove the penalization term (that is, $\alpha=0$), rouge scores drop a lot as the model cannot learn a reasonable attention without regularization. 
We will show qualitative cases for further explanation.
On the other hand, the performance will degrade with too high penalization weights, such as $\alpha=0.1$, as it will cause unstable training of contrastive loss.

\subsection{Qualitative Analyses}
We show qualitative results to demonstrate that our model can successfully extract salient sentences.
Table~\ref{tab:qualitative} gives two examples from CNN/Daily Mail test set.
Our extracted three sentences are marked in bold text, and corresponding equivalent summary sentences are marked with underlines.
We can see that DeepChannel can indeed find the most salient sentences from the document.
Besides, the redundant sentences are automatically avoided in our extractive results, which is attributed to the good property of the channel probability and our greedy strategy.

\begin{figure}[ht]
\includegraphics[width=\linewidth]{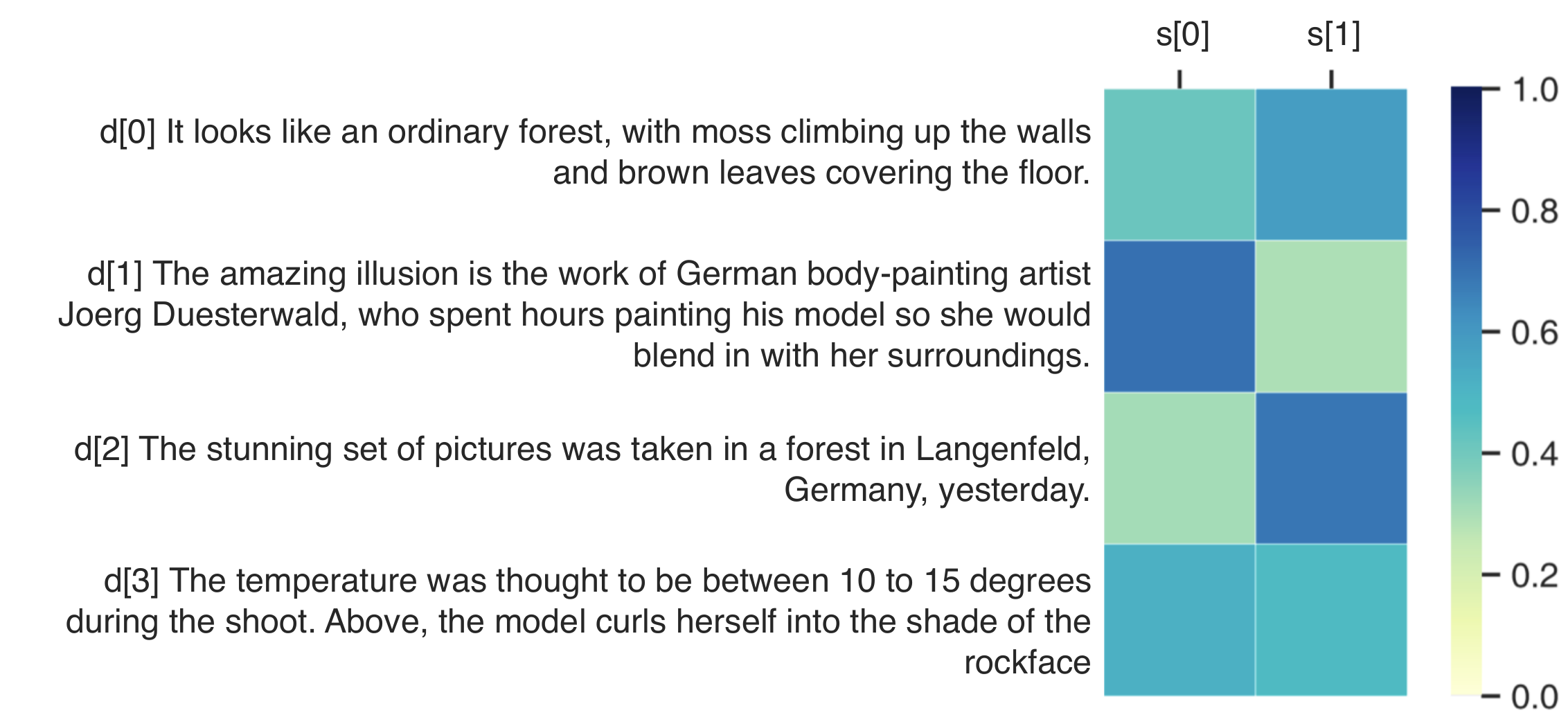}
\caption{Example of attention heatmap between document sentences (rows) and gold summary sentences (columns). s[0]: \textit{The illusion is the work of German body-painting artist Joerg Duesterwald, who spent hours painting his model}. s[1]: \textit{Stunning set of pictures was taken in front of a rockface in a forest in Langenfeld, Germany, yesterday}. Best viewed in color. }
\label{fig:goodattention}
\end{figure}

\begin{figure}[ht]
\includegraphics[width=\linewidth]{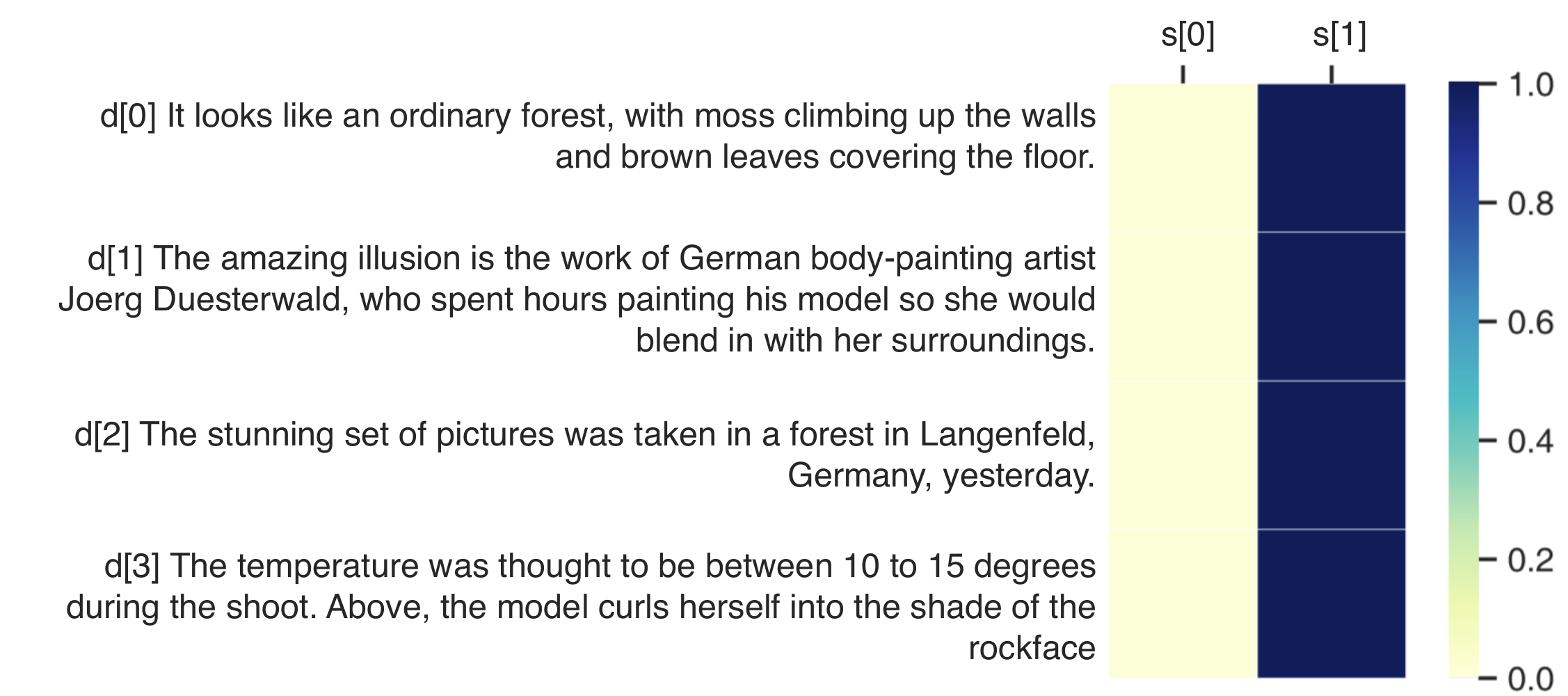}
\caption{Heatmap when removing the penalization term. We can see s[0] does not receive attention at all. Best viewed in color. }
\label{fig:badattention}
\end{figure}

Figure~\ref{fig:goodattention} shows an example of attention heatmap, where each row corresponds to a document sentence and each column corresponds to a sentence of the gold summary.
We can see that our model can successfully learn high attention scores for sentence pairs which have relevant semantics.

We also display the heatmap of the same document in the case of removing the penalization term during training (Figure~\ref{fig:badattention}).
We can see that all document sentences focus on s[1], while s[0] does not receive attention at all.
Our proposed penalization term can make sure that no summary sentence is left out.

\section{Conclusions and Future Work}
We propose DeepChannel, consisting of a deep neural network-based channel model and an iterative extraction strategy, for extractive document summarization.
Experiments on CNN/Daily Mail demonstrate that our model performs on par with state-of-the-art summarization systems.
Furthermore, DeepChannel has three significant advantages:
1) strong robustness to domain variations;
2) high data efficiency;
3) high interpretability.

For future work, we will consider more fine-grained, i.e., word-level, attention and extraction mechanisms.
Besides, we will try to take the language model $P(S)$ into account, to reflect the influence and coherence between adjacent sentences.

\section{Acknowledgments}
The work is supported by NSFC key projects (U1736204, 61533018, 61661146007), Ministry of Education and China Mobile Research Fund (No. 20181770250), and THUNUS NExT Co-Lab.

\fontsize{9.0pt}{10.0pt} \selectfont
\bibliography{reference_298}
\bibliographystyle{aaai}
\end{document}